\documentclass[11pt]{article} 
\usepackage{rldmsubmit, palatino}  
\usepackage[hyphens]{url}  
\usepackage{graphicx} 
\usepackage{amsmath}
\usepackage{wrapfig}

\usepackage{subfig}

\usepackage{caption} 
\DeclareCaptionStyle{ruled}{labelfont=normalfont,labelsep=colon,strut=off}

\usepackage{bbm}
\usepackage[round]{natbib}
\title{Designing Rewards for Fast Learning}
\author{
    Henry Sowerby \\
Department of Computer Science \\
Brown University \\
Providence, RI 02912 \\
    \texttt{henry\_sowerby@brown.edu} \\
    \And
    Zhiyuan Zhou \\
Department of Computer Science \\
Brown University \\
Providence, RI 02912 \\
    \texttt{zhouzy@brown.edu} \\
    \And 
    Michael L.\ Littman \\
Department of Computer Science \\
Brown University \\
Providence, RI 02912 \\
    \texttt{mlittman@cs.brown.edu} \\
}
\newcommand{\namecite}[1]{\citeauthor{#1}~(\citeyear{#1})} 
\begin{document}

\maketitle

\begin{abstract}
To convey desired behavior to a Reinforcement Learning (RL) agent, a designer must choose a reward function for the environment, arguably the most important knob designers have in interacting with RL agents. Although many reward functions induce the same optimal behavior~\citep{ng99}, in practice, some of them result in faster learning than others. In this paper, we look at how reward-design choices impact learning speed and seek to identify principles of good reward design that quickly induce target behavior. This reward-identification problem is framed as an optimization problem: Firstly, we advocate choosing state-based rewards that maximize the action gap, making optimal actions easy to distinguish from suboptimal ones. Secondly, we propose minimizing a measure of the horizon, something we call the ``subjective discount'', over which rewards need to be optimized to encourage agents to make optimal decisions with less lookahead. To solve this optimization problem, we propose a linear-programming based algorithm that efficiently finds a reward function that maximizes action gap and minimizes subjective discount. We test the rewards generated with the algorithm in tabular environments with Q-Learning, and empirically show they lead to faster learning. Although we only focus on Q-Learning because it is perhaps the simplest and most well-understood RL algorithm, preliminary results with R-max~\citep{brafman00} suggest our results are much more general. Our experiments support three principles of reward design: 1) consistent with existing results, penalizing each step taken induces faster learning than rewarding the goal. 2) When rewarding subgoals along the target trajectory, rewards should gradually increase as the goal gets closer. 3) Dense reward that's nonzero on every state is only good if designed carefully.
\end{abstract}


\acknowledgements{We are grateful to David Abel and his reward research group at Google DeepMind for discussions and feedback, as well as NSF, DARPA, and ONR for funding support.}

\startmain 

\section*{Problem Setting}
 


We formulate the RL problem as a Markov Decision Process, modeling the environment in terms of states $S$, actions $A$, reward function $R(s)$, discount factor $\gamma$, and transition function $T(s, a, s')$. Reward functions are often defined on state--action pairs~\citep{Puterman94}, but we limit ourselves to the simpler case in which a reward function depends only on the states~\citep{russell94}. To further simplify reward specification, states are described in terms of a vector of features $F(s,i)$ and rewards are computed as a linear combination of these features~\citep{ng00}. We seek to encourage, via rewards on state features, a target policy $\pi^+$~\citep{abel21} mapping states to actions. An optimal policy $\pi^*_R$ is one that maximizes the cumulative discounted expected reward from all states. We say a reward function $R$ is \emph{correct} if $\pi^*_R = \pi^+$.

To evaluate the ease of learning for a given correct reward function, we count the number of discrete timesteps where the learner's preference matches the target policy. We stray away from the usual benchmark of cumulative reward achieved
because reward is misleading in our case; designing a reward function that has across-the-board higher reward is not necessarily better at the task of encouraging the target behavior. 
Specifically, we define
$$\textrm{correct actions} = \sum_{t=1}^{T} \mathbbm{1}[ \mathrm{argmax}_{a} Q(s_t, a) = \pi^+(s_t)].$$

\begin{figure}[htbp!]
  \centering
  \begin{minipage}[b]{0.25\textwidth}
    \includegraphics[width=\textwidth]{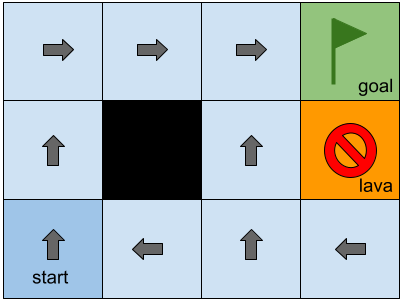}
    \caption{
    Russell/Norvig grid: The agent tries to get to the goal while avoiding the lava. At each step, the agent can choose to move in four cardinal directions, each of which has probability $0.8$ to transition in that direction, and probability $0.1$ to slip into the two orthogonal directions. The target policy is illustrated with arrows.
    }
    \label{f:rngrid}
  \end{minipage}
  \hfill
  \begin{minipage}[b]{0.35\textwidth}
    \includegraphics[width=\textwidth]{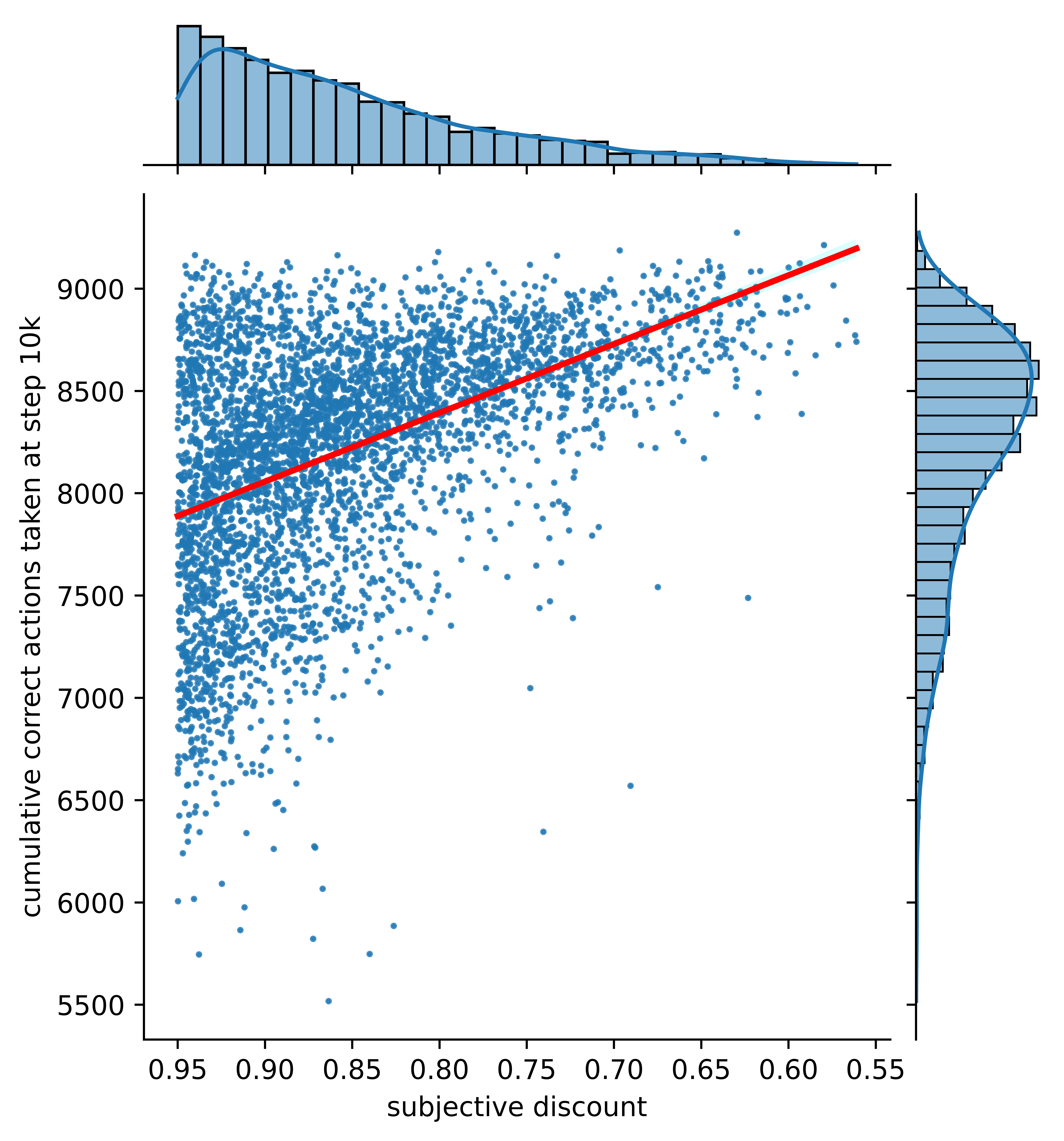}
    \caption{
    Cumulative correct actions of random reward functions vs. their subjective discount, Russell/Norvig grid, averaged over $1,000$ runs. Regression in red.
    }
    \label{fig:rn-random-vs-lp}
  \end{minipage}
  \hfill
  \begin{minipage}[b]{0.35\textwidth}
    \includegraphics[width=\textwidth]{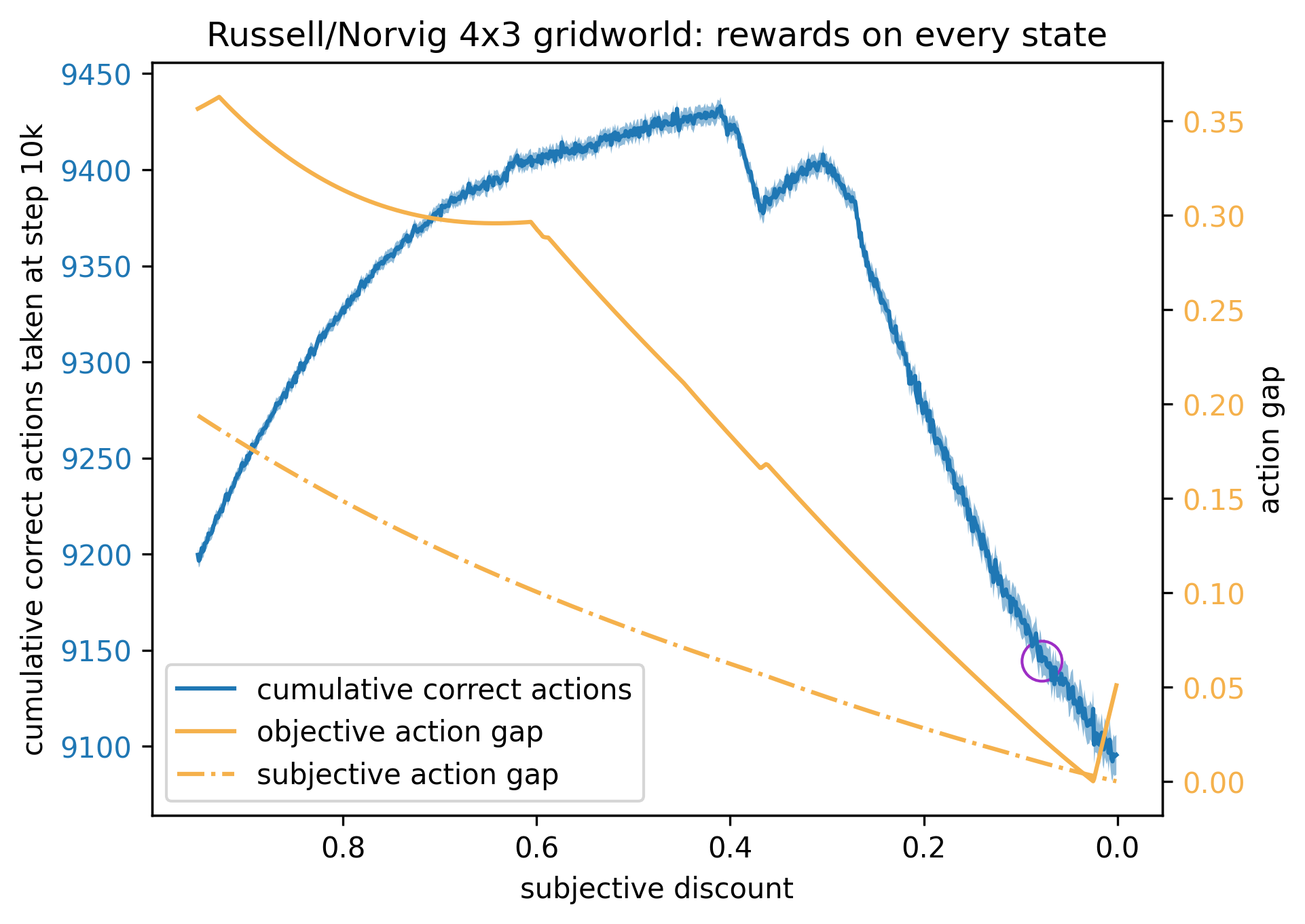}
    \caption{Learning performance (with 99\% confidence interval averaged over 5,000 runs) of linear-program-designed rewards and their action gap. The objective action gap is computed at $\gamma=0.95$, and the subjective action gap at $\tilde{\gamma}$. The purple circle marks the cumulative correct actions at threshold $\tilde{\delta} = 0.01$.
    }
    \label{fig:rn-performance-and-delta-vs-gamma}
  \end{minipage}
\end{figure}

We start by examining the diversity of correct reward functions for a given target policy. Consider the grid world example from \namecite{russell94}, described in Figure~\ref{f:rngrid}. To assess desirable aspects of different reward functions, we sample a collection of random reward functions that set $R(s) \in [-1, 1]$ for each state $s \in S$. Of the approximately $10,000,000$ randomly-generated reward functions, $5,000$ ($0.05\%$) were correct, and we ran Q-Learning on these reward functions for 10,000 steps. The right margin of Figure~\ref{fig:rn-random-vs-lp} shows the distribution of cumulative number of correct actions these rewards induce. The distribution shows that although each of these reward functions produce the target policy when fully optimized, they differ significantly in the speed with which Q-Learning aligns itself with this target policy. 


\subsection*{Subjective Discount}

In trying to understand the properties of reward functions that lead to faster learning, we leverage the following insight: The convergence of algorithms like Q-Learning and R-Max and even value iteration depends on the discount factor. Smaller discount factors lead to faster convergence.

In our setting, we cannot simply adopt a smaller discount factor because we consider it to be part of the environment. As such, we adopt an idea from \namecite{jiang15} and differentiate between two kinds of discount factors: the objective discount $\gamma$ of the environment and a \emph{subjective discount} $\tilde{\gamma}$, the latter of which is a property of the reward function. $\gamma$ is used by the Q-Learning agent to estimate returns during learning, and $\tilde{\gamma}$ to design rewards. 


Given a reward function $R$, we define $\pi^\gamma_R$ to be the policy of an agent optimizing $R$ in an environment with discount $\gamma$. The subjective discount for reward function $R$ and target policy $\pi^+$ is the smallest $\tilde{\gamma}$ such that
$\pi^{{\gamma}'}_R = \pi^+, \forall \gamma' \in [\tilde{\gamma},\gamma].$
That is, we want to know how small the discount factor could get while still encouraging the target policy. In our implementation, we use a binary search to efficiently compute the subjective discount.

Even though the learning agent does not have direct access to $\tilde{\gamma}$, it still seems to be a useful marker for good reward functions. Returning to Figure~\ref{fig:rn-random-vs-lp}, the x-axis plots the subjective discount against the cumulative number of correct actions that match our target policy. Although the two measures are not perfectly correlated, low subjective discounts are reliable estimators of fast learning. 

\subsection*{Reward Construction via Linear Programming}

To aid our exploration of the relationship between subjective discount and rewards for fast learning, we created an efficient algorithm for identifying correct reward functions with minimal subjective discounts. Whereas our random reward-function search was able to identify reward functions with subjective discounts of around $0.55$ for the Russell/ Norvig grid, our linear-programming-based optimization process can push this value down to effectively zero. The generated reward for step cost, goal reward, and lava penalty, respectively, was  $(-0.0223, +0.6119, -1)$, which outperforms the original rewards $(-0.04, +1, -1)$, with 88,364 vs. 83,666 correct actions, after 100k steps, averaged over 100 runs.

We make use of the discounted expected feature expectation~\citep{abbeel04} $D(s, i)$ and state--action feature expectations $D_a(s, i)$ for state $s$, action $a$, and feature $i$. They are defined using the feature value $F(s, i)$ and target policy $\pi^+$ at discount $\gamma$:
$$D(s, i) = F(s, i) + \gamma \sum_{s'} T(s,\pi^+(s),s') \cdot D(s', i)$$
$$D_a(s, i) = F(s, i) + \gamma \sum_{s'} T(s,a,s') \cdot D(s', i) $$
Similarly, we also define the same quantities $\widetilde{D}(i, s)$ and $\widetilde{D}_a(s, i)$ using subjective discount $\tilde{\gamma}$. It has been shown~\citep{syed} that the state value function and Q-value of the target policy can be expressed as $V^{\pi^+}(s) = \sum_{i} D(s, i) \cdot R(i)$ and $Q^{\pi^+}(s, a) = \sum_{i} D_a(s, i) \cdot R(i)$, where $R(i)$ is the reward function based on the feature $i$.

The goal of our optimizer is to construct a reward function that induces the target policy $\pi^+$ at the objective discount \emph{and} the lowest possible subjective discount. Our linear program is then formulated as:
\begin{alignat}{2}
  & \text{choose } \delta, & & R(i) \text{ to maximize }     \delta \nonumber \\
  & \text{subject to}    &       & \sum_{i} D(s, i) \cdot R(i) \geq \sum_{i} D_a(s, i) \cdot R(i) + \delta\nonumber\\
  &                &          & \sum_{i} \widetilde{D}(s, i) \cdot R(i) \geq \sum_{i} \widetilde{D}_a(s, i) \cdot R(i) + \delta \nonumber\\
  &                &          &  -1 \leq R(i) \leq 1           \nonumber\\
  &  &    & \forall \medspace i, s \in S, a \in A\setminus\{\pi^+(s)\} \nonumber
\end{alignat}

In words, choose the reward function such that, for each state, the target policy's action in that state is at least $\delta$ better than any other action (for both the objective and subjective discount factors). The role of the $\delta$ variable here
is
to make sure that the target policy's action is truly better ($\delta>0$) than the other actions. 


Next, we more closely evaluate the rewards constructed. For a range of $\tilde{\gamma} \in [0, 1]$, we ran the linear program to construct a range of reward functions, one for each $\tilde{\gamma}$.
Figure~\ref{fig:rn-performance-and-delta-vs-gamma} plots the cumulative correct actions taken in $10,000$ steps for these rewards. 
The result shows that the best learning performance is achieved at an intermediate value of subjective discount, and we hypothesize that the reason is a kind of policy optimization regularization~\citep{jiang15}.

\subsection*{Action Gap}

Although the maximization objective $\delta$ was included as a mathematical convenience, it measures a property that has been recognized as significant in the literature. The difference between the value of the optimal action and the second best action is known as the \emph{action gap}~\citep{farahmad}.
It is commonly believed that a large action gap is beneficial for learning, mitigating effects of estimation errors~\citep{lehnert18aaai}, improving policies derived from Q-values~\citep{bellemare2015increasing}, and achieving faster convergence rates~\citep{farahmad}. To hedge against action gap being optimized to $0$ as subjective discount decreases, we add the constraint $\delta \geq 0.01$.

\section*{Principles of Good Reward Design}


\subsection*{Advice: Penalize Steps}

In a first-of-its kind result, \namecite{koenig93} analyzed the time complexity of Q-Learning in a deterministic goal-based task, where they considered two two-featured correct reward functions: $(1,0)$ and $(0,-1)$. The first, the \emph{goal-reward} representation, provides the agent with a reward of $+1$ when the goal is reached and $0$ while en route. The second, the \emph{action-penalty} representation, penalizes the agent for each step taken, ending when the agent reaches the goal. 

From a zero-initialized Q-function, they argued that the time to learn a good behavior in the goal-reward representation can grow exponentially with the size of the state space, while in the action-penalty representation $O(|A||S|^2)$ steps suffice. Essentially, the zero-initialization and the zero rewards means the agent has no guidance and explores randomly until the goal is encountered by accident. With action penalties, the agent tries to avoid actions it has already taken, resulting in much more directed exploration.

We replicate this finding experimentally in a simple 60-state chain MDP
with objective discount $0.95$. It is a one-dimensional environment where the agent starts at the leftmost state, and needs to take rightward actions to arrive at the goal state on the opposite side. The two available actions transition the agent deterministically left or right. 

We compare the performance of the two reward functions just described with one that our linear program generates. For this environment, our linear program chooses $(1,-1)$, which we call ``combo'' as it combines the penalty and the reward.

At the objective discount of $0.95$, the action gaps for the three reward functions are approximately $0.0024$ (goal reward), $0.0485$ (action penalty), and $0.0509$ (combo). Since we use an action-gap threshold of $0.01$, goal reward doesn't have a well defined subjective discount, while action penalty achieves a subjective discount of $0.9249$, and combo $0.9238$, approximately. The subjective discounts suggest that combo and action penalty will be similar, and both better than goal reward. Figure~\ref{fig:comparing-reward-for-chain} confirms this hypothesis.


\subsection*{Advice: Reward Subgoals}

In the context of hierarchical reinforcement learning~\citep{parr97}, long-horizon tasks are broken into smaller tasks. Commonly, policies for the subtasks have their own pseudorewards that encourage subtask completion. In the context of a long-horizon goal, how should rewards be assigned to subtasks to encourage fast and accurate learning?

In naturally occurring problems, subtasks can break up goal-seeking sequences into heterogeneous chunks. In the simplified setting we study here, subtask completion states are spaced evenly along the 60-state chain and each has its own feature (and therefore can receive its own reward value).


\begin{figure}[hbt!]
  \centering
  \begin{tabular}{cc}
  \begin{minipage}[b]{0.48\textwidth}
    \includegraphics[width=\textwidth]{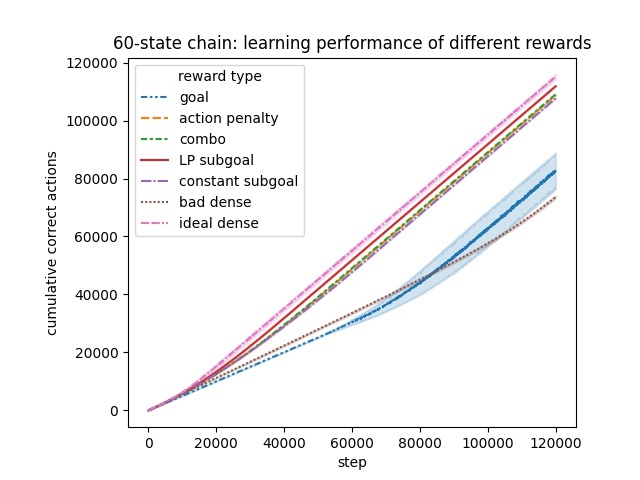}
    \caption{Comparison of reward functions on the 60-state chain.}
    \label{fig:comparing-reward-for-chain}
      \end{minipage}

&

         \begin{minipage}[b]{0.48\textwidth}
    \includegraphics[width=\textwidth]{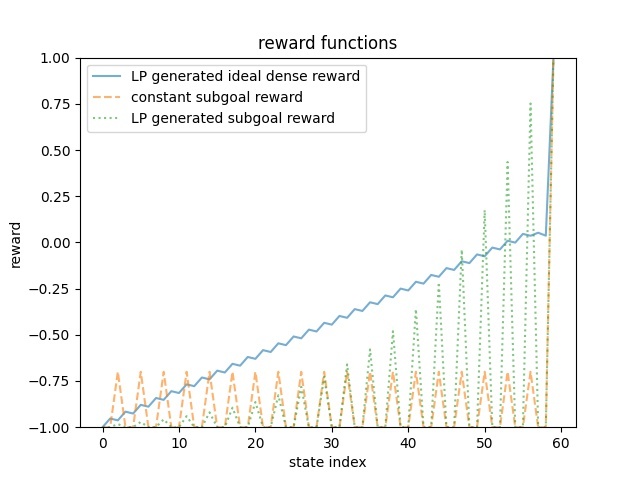}
    \caption{Two reward functions that place rewards on intermediate subgoals}
    \label{fig:chain-reward-shape}
  \end{minipage}
\end{tabular}

\end{figure}

We consider subgoal states placed every three states along the chain. The natural design choice would be to reward each of the resulting 19 subgoals the same reward value $r$, the goal $+1$, and all other states a step penalty $-1$, as suggested by the last section. However, it is not immediately obvious how to arrive at a value for $r$ such that the more frequently spaced rewards induce the target policy. One example that we have found is $r=-0.7$, a small advantage over the $-1$ rewarded non-subgoal states.

We compared the constant subgoal reward above to a subgoal reward profile generated by our linear program.
The reward functions are shown in Figure~\ref{fig:chain-reward-shape}. The constant subgoal reward has a subjective discount of $0.9510$, while the subgoal profile has a much lower subjective discount of $0.8232$. The learning results are in Figure~\ref{fig:comparing-reward-for-chain}.

From Figure~\ref{fig:comparing-reward-for-chain}, we observe significant differences in learning time; the subgoal profile reward learns significantly faster than the constant subgoal reward, and, the goal-only ``combo'' reward from the last section is included, which actually learns faster than the constant subgoal reward. This result suggests that only appropriate subgoal rewards speed learning, while inappropriate ones may actually lead to slower learning.

We find that, when defining subgoal rewards, it helps to gradually increase rewards as the agent gets closer to the goal state. This design helps counteract the effect of discounting, but also continually spurs the agent forward, much like an annual salary raise is considered to be a good motivator in the commercial sector.

\subsection*{Advice: Use Dense Rewards}

It is widely recognized that the kind of sparse rewards that come from rewarding only the goal is inferior to ``dense'' rewards like the subgoal rewards discussed in the previous section~\citep{smart02}.

Here, we look at the limiting case where every state has its own reward value. Figure~\ref{fig:chain-reward-shape} shows the reward function found by the linear program. It has a subjective discount of $0.2128$. 
Organizing rewards in a jagged shape lets the reward function have the largest possible state-by-state increase while still remaining in the required $-1$ to $+1$ range. Notice that the intermediate rewards are all negative in value---that's to prevent the agent from finding the intermediate rewards more attractive than reaching the goal.

Figure~\ref{fig:comparing-reward-for-chain} compares the Q-Learning performance on this LP-produced dense reward to the spare rewards
from the previous section. We also included a second fully dense reward function selected to have a low action gap of near $0$, which performs much worse. An important lesson here is that sparse rewards are indeed difficult to learn from, but not all dense rewards are equally good.

{\small \bibliography{main}}

\end{document}